\documentclass{article}
\usepackage{amssymb}
\usepackage[preprint]{neurips_2025}
\usepackage[utf8]{inputenc} 
\usepackage[T1]{fontenc}    
\usepackage{hyperref}       
\usepackage{url}            
\usepackage{booktabs}       
\usepackage{amsfonts}       
\usepackage{nicefrac}       
\usepackage{microtype}      
\usepackage{xcolor}         
\usepackage{algorithm}
\usepackage{algpseudocode}
\usepackage{caption}      
\usepackage{subcaption}   
\usepackage{mdframed}
\usepackage{multirow}
\usepackage{multicol}
\usepackage{rotating}
\usepackage{tabularx}
\usepackage{adjustbox}
\usepackage{graphicx}
\usepackage{array}
\usepackage{ragged2e}  
\usepackage{siunitx}     
\usepackage{wrapfig} 
\usepackage{colortbl}
\newcolumntype{C}{>{\Centering\arraybackslash}X} 

\begin{document}

\title{OTPTO: Joint Product Selection and Inventory Optimization in Fresh E-commerce Front-End Warehouses}

\author{%
  {
  Zheming Zhang$^{1}$ \hspace{0.3cm}
  Yan Jiang$^{1}$\thanks{\ Equal contribution.}  \hspace{0.3cm}
  Qingshan Li$^{2*}$\hspace{0.3cm}
  Ai Han$^{1}$\thanks{\textsuperscript{} Corresponding author <hanai5@jd.com>.} \hspace{0.3cm}
  }
   \\
  \vspace{0.1cm}
  \\
  { \normalfont $^1$JD.COM, Beijing, China~
  $^2$University of Chinese Academy of Sciences, CAS
}
}

\maketitle

\begin{abstract}
In China's competitive fresh e-commerce market, optimizing operational strategies, especially inventory management in front-end warehouses, is key to enhance customer satisfaction and to gain a competitive edge. Front-end warehouses are placed in residential areas to ensure the timely delivery of fresh goods and are usually in small size. This brings the challenge of deciding which goods to stock and in what quantities, taking into account capacity constraints. To address this issue, traditional predict-then-optimize (PTO) methods that predict sales and then decide on inventory often don't align prediction with inventory goals, as well as fail to prioritize consumer satisfaction. This paper proposes a multi-task Optimize-then-Predict-then-Optimize (OTPTO) approach that jointly optimizes product selection and inventory management, aiming to increase consumer satisfaction by maximizing the full order fulfillment rate. Our method employs a 0-1 mixed integer programming model OM1 to determine historically optimal inventory levels, and then uses a product selection model PM1 and the stocking model PM2 for prediction. The combined results are further refined through a post-processing algorithm OM2. Experimental results from JD.com's 7Fresh platform demonstrate the robustness and significant advantages of our OTPTO method. Compared to the PTO approach, our OTPTO method substantially enhances the full order fulfillment rate by 4.34\% (a relative increase of 7.05\%) and narrows the gap to the optimal full order fulfillment rate by 5.27\%. These findings substantiate the efficacy of the OTPTO method in managing inventory at front-end warehouses of fresh e-commerce platforms and provide valuable insights for future research in this domain.
\end{abstract}


\section{Introduction}
The COVID-19 pandemic has transformed the shopping habits of Chinese customers, leading to a greater dependence on purchasing fresh goods online \cite{lu2022comparative,guo2022has}. This shift has significantly boosted the growth of China’s fresh e-commerce market \cite{iimr} but has also heightened competition among these platforms. To differentiate themselves, fresh e-commerce platforms must continuously refine their operational strategies to enhance consumer satisfaction. A critical component of these strategies is the optimization of inventory management in front-end warehouses.

\begin{figure}[htbp]
	\centering
	\includegraphics[width=1.0\columnwidth]{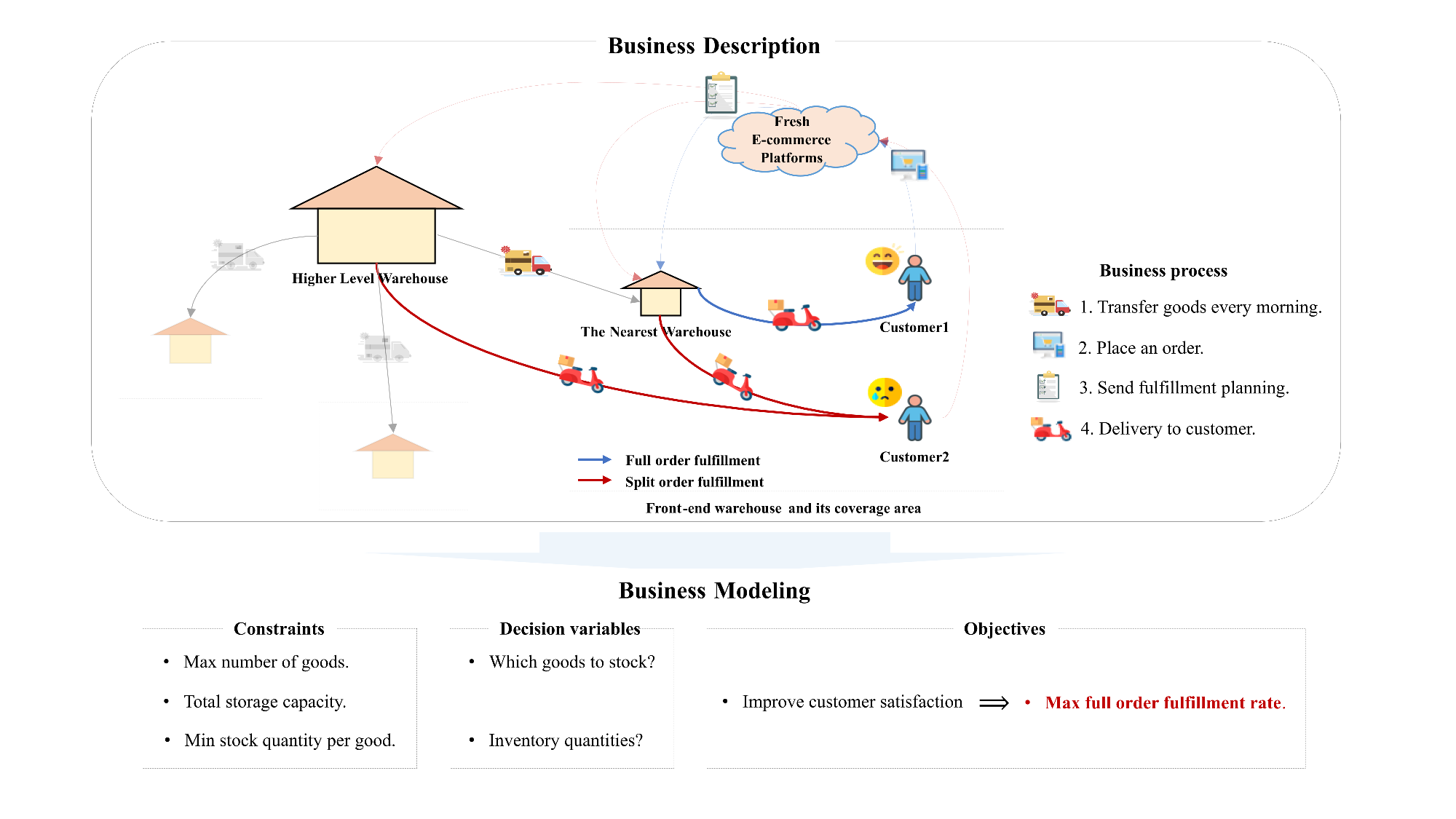}
        \vspace{-15pt}
	\caption{Overview of The Front-end Warehouse Model}
	\label{fig:intro-fig}
\end{figure}

Front-end warehouses, are typically located in residential areas within a 3 to 5-kilometer radius \cite{wan2022location} to ensure timely delivery of fresh goods. Order fulfillment in front-end warehouses follows a specific process: customers place orders through apps like JD.com’s 7Fresh, and the system gives priority to select the front-end warehouse closest to the customer's delivery address for fulfillment. When an order can be completely fulfilled by a single front-end warehouse, it is called full order fulfillment. If there are any shortages that need to be supplemented by higher level warehouses further away, it results in split order fulfillment. This means that an order is delivered to the consumer in several times(See Figure \ref{fig:intro-fig}), greatly reducing customer satisfaction. To improve consumer satisfaction, maximizing the full order fulfillment rate is crucial. Full order fulfillment is closely related to the types and quantities of goods stored in the warehouse, which is the core of inventory management decision. In our scenario, the small size of front-end warehouses often results in limited storage capacity. As a result, not all predicted demand can be accommodated in the front-end warehouse, and any excess must be placed in larger, higher level warehouses. Therefore, the main challenge of front-end warehouse management is to determine which goods to stock and the corresponding inventory quantity based on demand forecasting and capacity constraints, aiming to maximize the full order fulfillment rate and thereby improve customer satisfaction. This is the focus of our paper.

Traditional inventory management problems are classified into single-period inventory problems, known as the Newsvendor Problem (NVP), and multi-period inventory problems, referred to as the Multi-Period Inventory Problem (MPIP), based on the number of sales cycles \cite{shenoy2018introduction}. Our study focuses on fresh goods, which require daily inventory checks and restocking, thus classifying them as NVP. The conventional approach firstly predicts daily sales in the area covered by the front-end warehouse. Subsequently, decisions regarding the types and quantities of inventory goods are made based on these predictions, adhering to the Predict-Then-Optimize (PTO) framework \cite{WOS:001101335500001}. Despite its popularity, this method may encounter issues such as cumulative errors across stages \cite{2023practqiical}, which can lead to suboptimal final decisions. This is primarily due to two reasons: (1) a misalignment between prediction phase goals (sales volume) and decision phase goals (inventory goods and quantities), where the most accurate demand prediction does not necessarily translate into optimal inventory decisions; and (2) incomplete information during the decision stage, where relying solely on prediction results may overlook critical information contained in the original data, thereby resulting in suboptimal optimization outcomes.

To address these challenges, our study introduces a novel approach that more effectively integrates prediction and optimization. Specifically, we propose a multi-task Optimize-then-Predict-then-Optimize (OTPTO) approach to jointly determine product selection and inventory quantities. In the first optimization phase, we formulate a 0-1 mixed-integer programming model OM1 that describes product selection, inventory decision, and order fulfillment processes. This model aims to increase consumer satisfaction by optimizing the full order fulfillment rate within practical inventory constraints. We then use a commercial solver to obtain the historically optimal results for product selection and inventory quantities. In prediction phase, we build two parallel sub-models, the product selection model PM1 and the stocking model PM2. PM1 is a binary classification model that employs a LightGBM algorithm to learn from the historical optimal product selections provided by OM1 and predict which goods need replenishment in the future. PM2, on the other hand, is a regression model that also utilizes a LightGBM algorithm to predict the optimal inventory levels of each good in stock, based on the historical optimal results from OM1. To enhance the accuracy of PM1 and PM2, we propose effective strategies focusing on three aspects: samples, labels, and features. For samples, we develop a differentiated sampling strategy to address the distinct roles and requirements of PM1 and PM2. For labels, we introduce label generation and label smoothing strategies to mitigate sample inconsistency issues, where samples with similar feature values might receive different labels due to the constraints of the OM1 model. For features, beyond the common features used in sales prediction scenario, we incorporate four types of high-quality features: decision-making features, sales prediction features, clustering features, and SKU (Stock-Keeping Unit)-order cross-features. In the second optimization phase, we combine the PM1 and PM2 results and get the final result through a post-processing algorithm OM2. Following the OTPTO method, we conduct numerical experiments using real sales data from 7Fresh, a subsidiary of JD.com, to validate our approach. The experimental results demonstrate that our OTPTO method significantly increases the full order fulfillment rate by 4.34\% (relative to 7.05\%) and reduces the gap to the optimal full order fulfillment rate by 5.27\%. We also assess the individual contributions of different sample, label, and feature strategies through ablation experiments. Additionally, we validate the robustness of our method by evaluating its performance on datasets from other five front-end warehouses. The key innovations of our study can be summarized as follows:
\begin{itemize}
\item To the best of our knowledge, this is the first study to address the joint product selection and inventory decision problem in the context of fresh food e-commerce, focusing on front-end warehouses to increase consumer satisfaction by maximizing full order fulfillment rates, considering storage capacity constraints, product type constraints, and stock quantity constraints for each good.
\item Through numerical experiments, we demonstrate that the proposed OTPTO method outperforms the traditional PTO method. Additionally, the label strategy effectively mitigates issues of sample inconsistency caused by the constraints of the optimization problems, providing valuable insights for similar research.
\item We conduct experiments on real-world datasets. The results show the effectiveness and robustness of the OTPTO method, proving its great potential for widespread practical application.
\end{itemize}

\section{Related Work}
In this section, we first provide an overview of relevant research on fresh product inventory management. Following this, we delve into studies centered on the PTO method, examining its application in the field of inventory management.

\textbf{Newsvendor Problem and Inventory Management of Perishable Goods} 
This study addresses the inventory management challenges faced by front-end warehouses, which are essentially single-period inventory management problems, commonly referred to as the newsvendor problem. Under the assumption of uncertain demand, the newsvendor model requires decision-makers to determine the order quantity for each sales period. The objective is to maximize expected profit or minimize expected costs, including costs associated with spoilage and stockout \cite{shenoy2018introduction}. Since the model's introduction, researchers have conducted extensive studies \cite{mu2019multi}, considering factors such as capacity constraints \cite{erlebacher2000optimal}, budget constraints \cite{zhou2015multi,zhang2020analysis}, and decision-makers’ risk preferences \cite{wu2013risk}. In recent years, the feature-based newsvendor model, which leverages data-driven optimization, has become a prominent research topic. This approach not only considers historical demand data but also incorporates contextual information such as product features, weather conditions, and the day of the week \cite{zhang2024optimal,olivares2024constructing,serrano2024703}. 
Our research distinguishes itself from existing studies by emphasizing the crucial connection between fresh front-end warehouse inventory and the order fulfillment process, which is essential for enhancing customer experience. Our method integrates various constraints and contextual information, providing a novel perspective and solution for inventory management. 

\textbf{PTO method} In practical scenarios, many problems can be effectively addressed using both prediction and optimization techniques. Consequently, the Predict-Then-Optimize (PTO) method has long been widely adopted across various industries, including healthcare \cite{WOS:000312468800003,WOS:000337721400015}, business analytics \cite{WOS:000338352900012,WOS:000367776900002}, supply chain management \cite{WOS:000352821400003,WOS:000228113800002}, e-commerce \cite{WOS:000375601500006,WOS:000457325000015}, manufacturing \cite{WOS:000290752200001,WOS:000423271600004}, and inventory management \cite{jia2024scenario}. Despite the popularity of the PTO approach, it also has drawbacks. One of the primary issues is the misalignment between the goals of its prediction and optimization stages. This misalignment arises because the prediction model forecasts parameters for the decision model, while the optimization stage uses these predicted parameters to determine the optimal decision variables. This process can result in information loss and error propagation. To mitigate this issue, recent advances have introduced end-to-end PTO frameworks such as “Smart PTO”  \cite{WOS:001101335500001} and “PyEPO” \cite{tang2023pyepo}. These methods integrate the optimization model into the prediction model as a loss function, enabling end-to-end optimization, which reduces information loss and error propagation. However, this approach requires the prediction model to accurately output all the unknown parameters needed by the decision model simultaneously, making it extremely difficult to train. These requirements render the method unsuitable for large-scale problems, thereby limiting its practical applicability. To address these issues, our proposed OTPTO method first identifies historically optimal decisions. During the prediction phase, it directly learns the values of the optimal decision variables instead of the parameters of the decision model. Finally, the predicted results are fine-tuned to meet the constraints of the decision model. This results in more accurate and reliable outcomes.

\section{Problem Definition}
\label{sec:pd}
In the context of fresh e-commerce retail, consumer satisfaction is primarily determined by three key factors: (1) the price and quality of the products, (2) the delivery efficiency, and (3) the likelihood of order splitting (as shown in Figure \ref{fig:intro-fig}). Our research concentrates on the third factor, which is directly related to consumer satisfaction \cite{yee2008impact,yan2015customer,rahiminezhad2022financial,hong2019analyzing,cui2023sustaining}. Therefore, our work primarily aims to enhance consumer satisfaction by maximizing the full order fulfillment rate. In terms of fulfillment costs, increasing the full order fulfillment rate helps reduce the likelihood of order splitting, which decreases the times of delivery, thus reducing fulfillment costs. Additionally, spoilage and stockout costs, which are usually considered in demand forecasting \cite{shenoy2018introduction}, are not the focus of this paper.

We define the following objective function that maximizes the average full order fulfillment rate over the future $T$ days:

\begin{equation}
	\max r = \frac{1}{T} \sum_{t=1}^T r_t = \frac{1}{T} \sum_{t=1}^T \frac{O\_F_t}{O_t}
\end{equation}

where $O_t$ and $O\_F_t$ denote the number of all orders and the number of fully-fulfilled orders respectively in a specified front-end warehouse on date $t$ (i.e. $t=0,1,\cdots,T$).

Besides, the inventory management of the front-end warehouse needs to meet the following three different types of capacity constraints:
\begin{enumerate}
    \item[(1)] Product types constraint: The maximum number of SKU types that can be stored in the front-end warehouse each day is $K$.
    \item[(2)] Storage capacity constraint: The maximum quantity of SKUs that can be stored in the front-end warehouse each day is $N$.
    \item[(3)] Stock quantity constraint: The minimum daily inventory requirement for each SKU is $B$.
\end{enumerate}

\section{Multi-Task OTPTO Pipeline}
\begin{figure}[htbp]
	\centering
	\includegraphics[width=1.0\columnwidth]{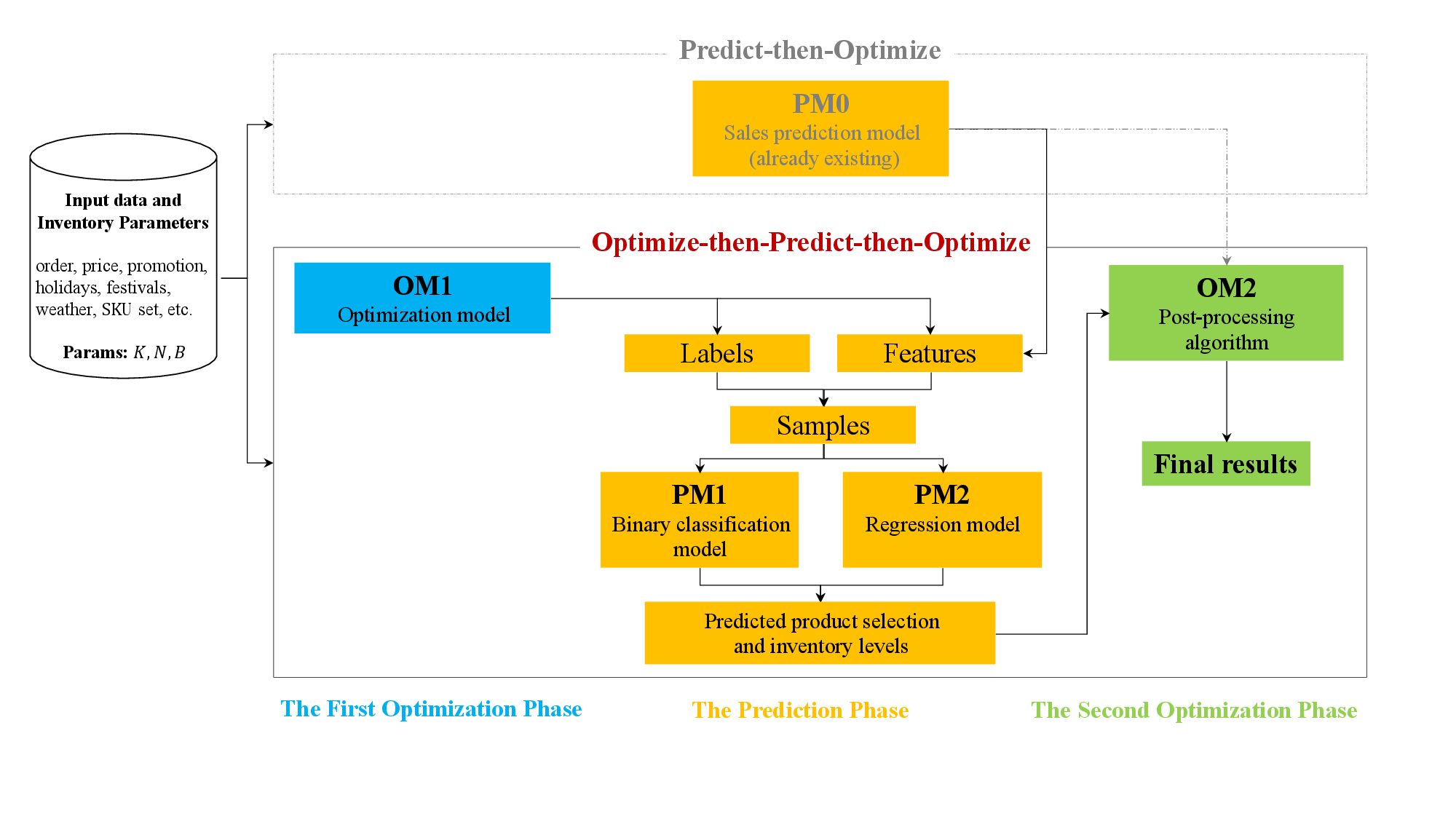}
        \vspace{-20pt}
	\caption{OTPTO Pipeline}
	\label{fig:OTPTO-pipeline}
\end{figure}

In this section, we introduce the OTPTO method, a multi-task approach for joint product selection and inventory optimization(see Figure \ref{fig:OTPTO-pipeline}). This method is systematically divided into three phases: the first optimization phase, the prediction phase and the second optimization phase. 

In the first optimization phase, we formulate a 0-1 mixed-integer programming model OM1 to describe the processes of product selection, inventory decision, and order fulfillment. We then utilize a commercial solver to obtain the historically optimal results for product selection and inventory quantity. During the prediction phase, the problem is decomposed into two parallel sub-problems. Independent models are constructed for the product selection prediction PM1 and the stocking quantity prediction PM2, both of which are trained using the results from OM1 and PM0 (already existing sales forecasting model). OM1 provides features and labels, while PM0 only provides features. Ultimately, in the second optimization phase, we merge the results of PM1 and PM2 and make final adjustments using a post-processing algorithm OM2. This process yields a product selection and stocking plan that meets the capacity constraints outlined in Section \ref{sec:pd}.

\subsection{The First Optimization Phase}
\label{sec:om}

In general, the stocking strategy actually implemented is often not optimal. Inspired by Qi et al \cite{2023practqiical}, we construct a joint optimization model for product selection and inventory quantity based on historical transaction data to obtain the training labels and inventory-related features of the prediction phase, that is, the optimal stocking results. For the independent problem concerning a specific front-end warehouse on a given sale day, we define: $O=\{1,2,\cdots, n\} $ as the set of order ids sorted according to their arrival time; $I=\{1,2,\cdots,m\}$ as the set of SKUs sold on the fresh e-commerce app; $ I_o$ represents the set of SKUs purchased by order $o$, with $I_o\subseteq I$; $d_i$ is the sales volume of SKU $i (i \in I)$; the continuous decision variable $x_i$ represents the stock quantity for SKU $i$; the binary decision variable $y_i$ indicates whether SKU $i$ is stocked; the binary decision variable $z_{oi}$ indicates whether SKU $i$ in order $o$ is supplied only by this warehouse; the binary decision variable $p_o$ indicates whether order $o$ is fully fulfilled by this warehouse; the statistic $c_{oi}$ represents the cumulative sales volume of SKU $i$ from order $1$ to order $o$; the statistic $s_o$ represents the number of SKUs purchased in order $o$; the constants $K,N,B$ are discussed before. $\delta$ and $M $ denote very small and very large constants, respectively.


\begin{equation}
\label{eq:obj1}
\max \frac{1}{n} \sum_{o \in O} p_o 
\end{equation}
\begin{equation}
\label{eq:k-c}
\sum_{i \in I} y_i \leq K
\end{equation}
\begin{equation}
\label{eq:n-c}
\sum_{i \in I} x_i \leq N
\end{equation}
\begin{equation}
\label{eq:b-c}
x_i \geq B y_i \quad \forall i \in I
\end{equation}
\begin{equation}
\label{eq:xub}
x_i \leq \max(B, d_i) \quad \forall i \in I
\end{equation}
\begin{equation}
\label{eq:rela-oi-1}
x_i - c_{oi} + \delta \leq M z_{oi} \quad \forall i \in I, o \in O
\end{equation}
\begin{equation}
\label{eq:rela-oi-2}
x_i - c_{oi} \geq M(z_{oi} - 1) \quad \forall i \in I, o \in O
\end{equation}
\begin{equation}
\label{eq: rela-os-1}
\sum_{i \in I_o} z_{oi} - s_o + \delta \leq Mp_o \quad \forall o \in O 
\end{equation}
\begin{equation}
\label{eq: rela-os-2}
\sum_{i \in I_o} z_{oi} - s_o \geq M(p_o - 1) \quad \forall o \in O 
\end{equation}
\begin{equation}
\label{eq:rela-xy-1}
x_i \leq My_i \quad \forall i \in I
\end{equation}
\begin{equation}
\label{eq:rela-xy-2}
M x_i \geq y_i \quad \forall i \in I 
\end{equation}
\begin{equation}
\label{eq:x_var}
x_i \geq 0 \quad \forall i \in I 
\end{equation}
\begin{equation}
\label{eq:y_var}
y_i \in \{0, 1\} \quad \forall i \in I 
\end{equation}
\begin{equation}
\label{eq:p_var}
p_i \in \{0, 1\} \quad \forall o \in O 
\end{equation}
\begin{equation}
\label{eq:z_var}
z_{oi} \in \{0, 1\} \quad \forall i \in I, o \in O 
\end{equation}

This model is a 0-1 mixed integer linear programming model. Expression (\ref{eq:obj1}) aims to maximize the full order fulfillment rate for the front-end warehouse; Expression (\ref{eq:k-c}) represents the constraints on the number of different SKUs stocked in the front-end warehouse; Expression (\ref{eq:n-c}) deals with the total storage capacity constraints of the front-end warehouse; Expression (\ref{eq:b-c}) sets the minimum inventory requirements for SKUs if $y_i = 1$(i.e., SKU $i$ is selected); Expression (\ref{eq:xub}) sets the maximum stock quantity of SKU $i$ based on sales volume. Expressions (\ref{eq:rela-oi-1},\ref{eq:rela-oi-2}) specify that for SKU $i$ within order $o$, if the inventory is sufficient, then SKU $i$ of order $o$ is supplied only by this warehouse, i.e., when $x_i\geq c_{oi}$, then $z_{oi}=1$; Expressions (\ref{eq: rela-os-1},\ref{eq: rela-os-2}) determine that an order is considered fully fulfilled if all SKUs within order $o$ are supplied only by this warehouse, i.e., when $\sum_{i\in I_o} z_{oi} = s_o$, then $p_o=1$; Expressions (\ref{eq:rela-xy-1},\ref{eq:rela-xy-2}) describe the relationship between decision variables $x_i$ and $y_i$, indicating that only selected SKUs can be stocked; Expression (\ref{eq:x_var},\ref{eq:y_var},\ref{eq:p_var},\ref{eq:z_var}) defines the range of values for the decision variables. The results are then obtained directly using a commercial solver.

\subsection{The Prediction Phase}
After the first optimization phase, we build the product selection prediction model PM1 and the stocking quantity prediction model PM2. The former is a binary classification model that predicts which products to select, while the latter is a regression model that predicts the amount of inventory. Both PM1 and PM2 use the robust and reliable LightGBM model \cite{grinsztajn2022tree,WOS:000452649403021} with the same features as input to directly learn the outcomes from OM1. Furthermore, we propose effective strategies focusing on samples, labels, and features to improve the accuracy of PM1 and PM2.

\subsubsection{Sample Strategy}
The minimum stock quantity constraint in Section \ref{sec:pd} results in a non-continuous distribution of the optimal stocking quantity, $x_i^\star$(for any given SKU $i$, either $ x_i^\star=0$ or $x_i^\star \geq B$), which makes the regression model PM2 difficult to learn. To mitigate the above issue, we exclude samples where $x_i^\star=0$. This further enhances the inventory forecasting ability of PM2, but weakens the ability of PM2 to predict product selection(if a SKU is not selected, $ x_i^\star=0$), which can be compensated by PM1.

\subsubsection{Label Strategy}
\begin{figure}[htbp]
	\centering
	\includegraphics[width=1.0\columnwidth]{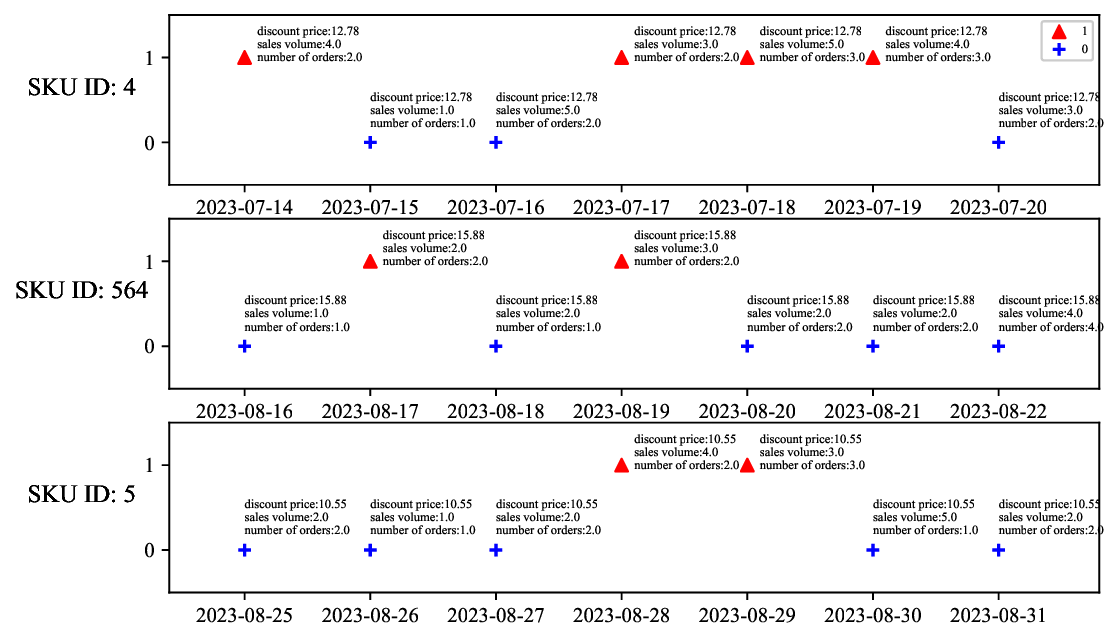}
	\caption{OM1's inventory decision of the same SKU across days. In cases where price, sales volume, and order quantity are all close, the same SKU may not be stocked every day in the OM1's optimal solution. Triangles for stocked SKUs, plus signs for non-stocked}
	\label{fig:stock-cross-days}
\end{figure}

Sample consistency is crucial for ensuring the effectiveness of machine learning algorithms. This suggests that samples with similar features should produce similar labels. By analyzing the decision results from OM1, the following two phenomena are observed:

\begin{enumerate}
	\item[(1)] On the same day, SKUs with similar sales performance have different optimal $y_i^\star$ values(see Table \ref{tab:s322-1});
	\item[(2)] For the same SKU, the sales volume of different days are similar, but the corresponding optimal $y_i^\star$ values are not the same, as shown in Figure \ref{fig:stock-cross-days}.
\end{enumerate}

\begin{figure}[htbp]
	\centering
	\includegraphics[width=1.0\columnwidth]{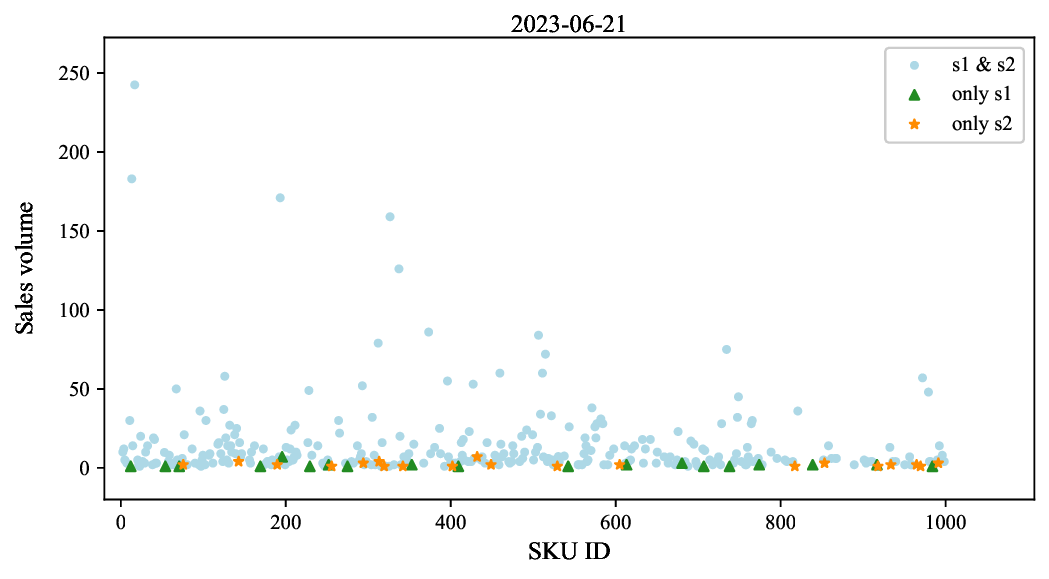}
	\caption{Product selection results from OM1 on 2023-06-21}
	\label{fig:stock-0621}
\end{figure}

We find there are many equivalent optimal solutions or near-optimal solutions in OM1 model. These solutions can achieve the same or similar full order fulfillment rates, but the SKU combinations are very different, contributing to the phenomenon 1. As shown in Figure \ref{fig:stock-0621}, s1 and s2 represent two different sets of equivalent optimal solutions on the same day. The blue points indicate SKUs shared by s1 and s2, which are usually high-volume products. The green and orange points represent SKUs unique to the s1 and s2, respectively, most of which are low-volume products. Practical experience shows that high-selling products help to cover more orders, and low-selling products help to match high-selling products to promote full order fulfillment. It can be seen that the participation of low-selling products in stocking plays a key role in improving the full order fulfillment rate. Given that OM1 solutions on different days are independent of each other, there are more obvious differences in SKU combinations across various days, resulting in phenomenon 2. To address the above issues, we implement two strategies to refine the $ y_i^\star$ :

(1) $y_i^\star$ \textbf{Generation}: We create a sub-objective to maximize the total Gross Merchandise Value (GMV) of all fully-fulfilled orders. Then, the solution that guarantees both the highest full order fulfillment rate and the largest total GMV is obtained from the set of equivalent optimal solutions. This method ensures the uniqueness of the OM1 solutions on the same day and makes these solutions on different days in the same distribution as much as possible, so as to reduce the fluctuation of $ y_i^\star$.
\begin{equation}
\label{eq:om-obj1}
\max \frac{1}{n} \sum_{o\in O} p_o
\end{equation}
\begin{equation}
\label{eq:om-obj2}
\max \sum_{o\in O} gmv_o p_o 
\end{equation}
where $gmv_o$ represents the GMV of order $o$. This multi-objective optimization problem can be solved by linearly weighting two objectives into a single objective:

\begin{equation}
\label{eq:om-obj1&obj2}
\max \frac{1}{n}( \sum_{o\in O} p_o + \sum_{o\in O} \frac{gmv_o}{\sum_{o\in O} gmv_o} p_o) 
\end{equation}

(2) $y_i^\star$ \textbf{Smoothing}: To tackle phenomenon 1, we apply the K-Means algorithm to cluster all SKUs on the same day and adjust the $y_i^\star$ based on the clustering results, calling this process “cross-sectional smoothing”. For phenomenon 2, we modify $ y_i^\star$ according to the proportion of historical stocking days to active sales days (i.e., the sales volume is greater than zero) for each SKU, which we refer to as “time series smoothing”. For details, see Algorithm \ref{alg:Algo1} in Appendix \ref{alg:Algorithms}.

\begin{table}[t]
    \centering
    \caption{Stocking status of SKUs with similar sales performance on the same day}
    \label{tab:s322-1}
    \begin{tabular}{ccccccc}
    \toprule
        \textbf{Date} & \textbf{sku\_id} & \textbf{$y_i^\star$} & \textbf{prc\_disct} & \textbf{sale\_qtty} & \textbf{ord\_cnt} \\ \midrule
        2023-08-06 & 2   & \textbf{0} & 18.58 & 4 & 4  \\ 
        2023-08-06 & 280 & \textbf{1} & 18.80 & 4 & 4  \\ 
        2023-08-19 & 51  & \textbf{0} & 30.38 & 3 & 2  \\ 
        2023-08-19 & 904 & \textbf{1} & 30.38 & 3 & 2  \\ 
        2023-08-20 & 754 & \textbf{0} & 13.56 & 8 & 5  \\ 
        2023-08-20 & 28  & \textbf{1} & 13.68 & 10 & 5  \\ 
        2023-08-20 & 29  & \textbf{1} & 13.68 & 7 & 4  \\ \bottomrule
    \end{tabular}
\end{table}

\subsubsection{Feature Strategy}
In the feature engineering stage, we augment the PM1 and PM2 model
with four types of features:

\textbf{Decision-making Features} These primarily include statistics on historical optimal inventory levels, historical inventory days, the number and the GMV of orders fulfilled.

\setlength{\intextsep}{4pt}           
\setlength{\belowcaptionskip}{20pt}   

\begin{wrapfigure}{r}{0.35\textwidth}
  \centering
  \vspace{-10pt}
  \includegraphics[width=0.36\textwidth]{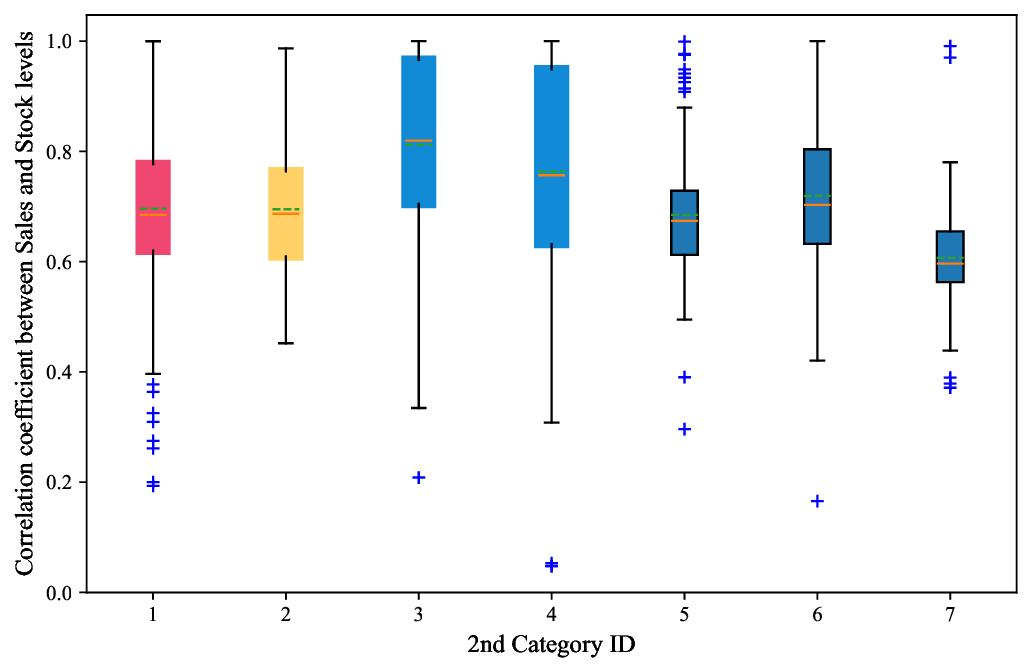}
  \vspace{-20pt}
  \caption{Correlation between sale quantity and optimal inventory quantity across second categories}
  \label{fig:corr-ss}
  \vspace{-20pt}
\end{wrapfigure}


\textbf{Sales Prediction Features} The results of the OM1 model show a strong correlation between the optimal inventory levels $x_i^\star$ and the actual sales on the corresponding day, with the median correlation coefficients exceeding 0.6 across all second categories (see Figure \ref{fig:corr-ss}). Consequently, we incorporate sales prediction features into our model, including sales forecast values, the mean of sales forecast residuals, and the standard deviation of sales forecast residuals, all coming from the already existing sales forecasting model PM0.



\setlength{\intextsep}{4pt}           
\setlength{\belowcaptionskip}{20pt}   

\begin{wrapfigure}{r}{0.35\textwidth}
  \centering
  \vspace{-10pt}
  \includegraphics[width=0.36\textwidth]{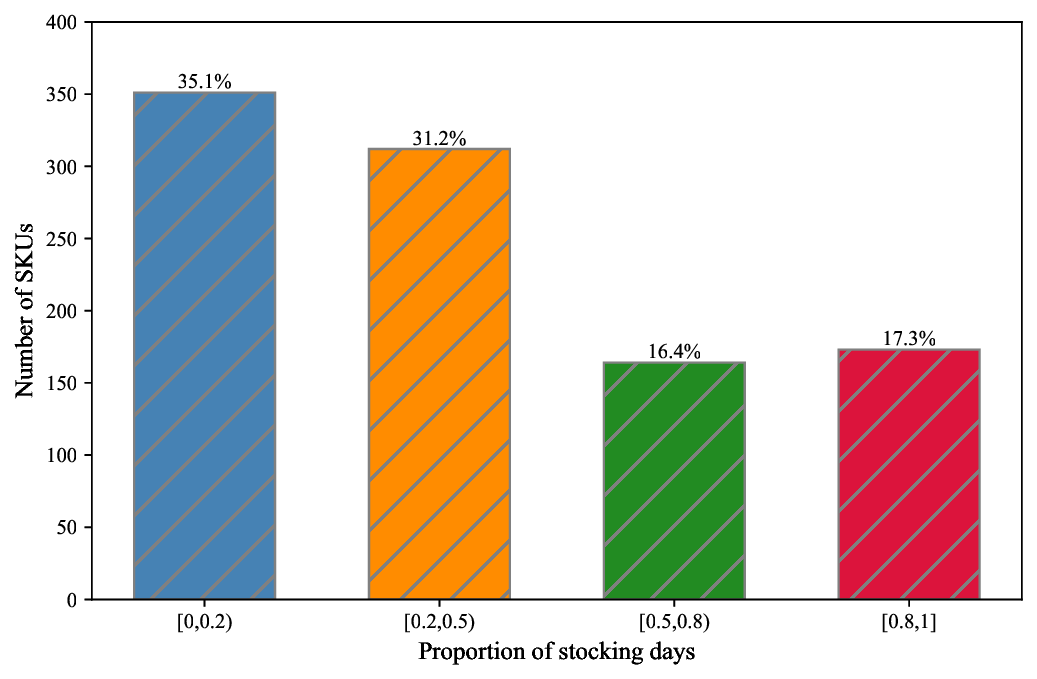}
  \vspace{-20pt}
  \caption{Distribution of Stocking Frequency}
  \label{fig:dist-stock}
  \vspace{-20pt}
\end{wrapfigure}

\textbf{Clustering Features} In Figure \ref{fig:dist-stock}, 17.3\% of SKUs are frequently stocked, while 35.1\% are seldom stocked. To differentiate SKUs based on stocking frequency, related features are introduced. Using the proportion of historical stocking days (total stocking days/total sales days) and the historical average daily inventory quantity (total inventory quantity/total sales days) as attributes, we employ the K-Means algorithm to classify SKUs into $\rho$ 
clusters, incorporating these clustering features into our predictive model.

\begin{figure}[htbp]
	\centering
	\includegraphics[width=1.0\columnwidth]{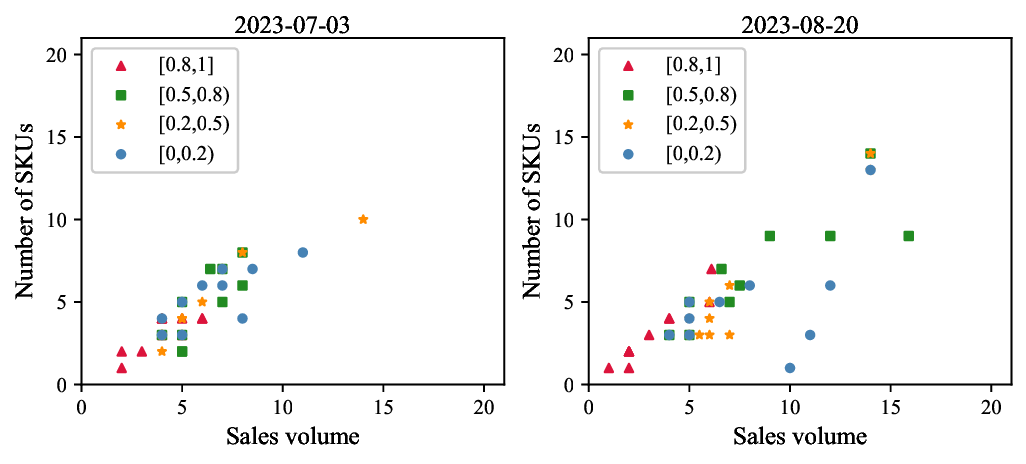}
	\caption{Sales and types in orders. The points in the graph represent SKUs, with the axes representing the average sales volume and average SKUs number of the order sets associated with those SKUs}
	\label{fig:sale-cate}
\end{figure}

\textbf{SKU-Order Cross-Features} Through the comparative analysis of OM1 outcomes across various days, frequently stocked SKUs are typically: a) appear in a larger number of orders; b) related to orders that contains only a small number of SKUs and low sales volume(see Figure \ref{fig:sale-cate}, red triangle points represent SKUs with a proportion of stoking days in [0.8,1]). This phenomenon is well explained: the former helps to cover more orders, while the latter facilitates complete the full order fulfillment. Therefore, we also integrate SKU-order cross-features into our model, which includes statistics on quantities, sales, and product types in historical orders covered by SKUs.

\subsection{The Second Optimization Phase}
After the above phases, the prediction results of PM1 and PM2 are obtained. However, these results may not satisfy the three capacity constraints mentioned in Section \ref{sec:pd}. We propose a post-processing algorithm OM2 to adjust the forecast results. For details, see Algorithm \ref{alg:Algo2} in Appendix \ref{alg:Algorithms}.

\section{Numerical Experiment}
In this part, we evaluate the performance of our proposed OTPTO method using real transaction data from a 7Fresh supermarket, a subsidiary of JD.com. Specifically, we design three experiments: (1) a comparative analysis between OTPTO and PTO method to determine the superiority of OTPTO; (2) an ablation experiment to evaluate the impact of different samples, labels, and feature strategies on the OTPTO method individually; and (3) a robustness analysis by testing the performance on datasets from the other five front-end warehouses.

\textbf{Dataset} The dataset covers a period of four months (from 2023-06 to 2023-09) and includes six front-end warehouses, each with an average daily order volume exceeding 1,000. It encompasses over 1,000 SKU types, across seven secondary categories. We utilize various data attributes, including SKU, brand, category, price, promotions, sales, orders, front-end warehouses, holidays, and weather, etc. These attributes are aggregated across different dimensions, such as SKUs, brands, categories, warehouses, and time periods (i.e., year, month, week, and day), to create new features. The already existing PM0 model forecasts sales based on these features. Our proposed OM1/PM1/PM2/OM2 models also use these raw and aggregated data as inputs.

\textbf{Parameter} The parameters for our study are as follows: The training dataset consists of historical data from a three-month period (2023-06-01 to 2023-08-31), and the test dataset includes data from the following week (2023-09-01 to 2023-09-07, with $T=7$ days). The upper limit of daily SKU stocking types is $K=350$, the upper limit of daily total stocking quantity is $N=9000$, and the minimum daily stocking quantity for each SKU is $B=10$. The hyperparameter values determined after search are: feature optimization clustering parameter $\rho=4$, label optimization clustering parameter $\lambda=80$, cross section smoothing threshold $\mu=0.8$, time series smoothing threshold $\gamma=0.8$, OM1 model constants $\delta=\mathrm{1e-3}$, $M=\mathrm{1e5}$. For details about parameters of PM1 and PM2, see Table \ref{tab:PM1-PM2-param} in Appendix \ref{tab:Tables}.

\textbf{PTO method} To compare the effects, we also implement the PTO method, first sort the SKUs in descending order according to the predicted sales volume, and then perform a improved greedy search algorithm to make product selection and inventory quantity decisions, see Algorithm \ref{alg:Algo2} in Appendix \ref{alg:Algorithms}.

\subsection{Results}
As can be seen from Table \ref{tab:OTPTO-pto-re}, the OTPTO method has obvious advantages over the PTO method. The full order fulfillment rate in 7 days has increased by 4.34\% (relative to 7.05\%). The single-day full order fulfillment rate has increased the most on the first day, reaching 5.5\%, and the minimum increase is 3.56\%. Moreover, the gap from the optimal solution(OPT) has narrowed significantly, from the original 25.29\% (1-61.57\%/82.41\%) to 20.02\%(1-65.91\%/82.41\%), an absolute decrease of 5.27\%. 

\begin{table}[!ht]
    \centering
    \caption{Results of OTPTO and PTO}
    \label{tab:OTPTO-pto-re}
    \renewcommand{\arraystretch}{1.2}
    \begin{tabular}{ccccccccc}
    \toprule
        \textbf{Date} & \textbf{Ord qtty} &  \multicolumn{3}{c}{\textbf{Full ord fulfillment rate}}  & \textbf{} \\ \hline
        ~ & ~ &  \textbf{OTPTO} & \textbf{PTO} & \textbf{OPT} & \textbf{Diff} \\ \cline{3-6}
        2023-09-01 & 1126 &  72.11\% & 66.61\% & 83.92\% & 5.50\%  \\ 
        2023-09-02 & 1230 &  58.21\% & 54.07\% & 81.06\% & 4.14\%  \\ 
        2023-09-03 & 1320 &  64.02\% & 60.15\% & 81.97\% & 3.87\%  \\ 
        2023-09-04 & 1137 &  66.14\% & 62.01\% & 82.06\% & 4.13\%  \\ 
        2023-09-05 & 1017 &  69.32\% & 64.21\% & 82.30\% & 5.11\%  \\ 
        2023-09-06 & 1171 &  61.91\% & 57.81\% & 82.41\% & 4.10\%  \\ 
        2023-09-07 & 1068 &  69.66\% & 66.10\% & 83.15\% & 3.56\%  \\ 
        \textbf{Avg} & \textbf{1153} & \textbf{65.91\%} & \textbf{61.57\%} & \textbf{82.41\%} & \textbf{4.34\%}  \\ 
        \bottomrule
    \end{tabular}
\end{table}

Besides, SKU combinations of the OTPTO and PTO methods differ greatly(as shown in Table \ref{tab:re-inv} in Appendix \ref{tab:Tables} and Figure \ref{fig:stock-0901}, the blue points indicate the OTPTO and PTO methods have shared SKUs, and the green and orange points represent the unique SKUs of OTPTO and PTO respectively). The OTPTO method has more low-selling products in stock, and these low-selling products are combined with high-selling products to promote higher full order fulfillment rate. The reason is that the stocking plan obtained by PTO method usually selects high-selling products for inventory, without considering the impact of low-selling products. On the contrary, the OTPTO method proposed in this article directly learns the historical optimal plan for the combination of high- and low-selling products, thereby improving the decision-making quality.

\begin{figure}[htbp]
	\centering
    \includegraphics[width=0.5\textwidth]{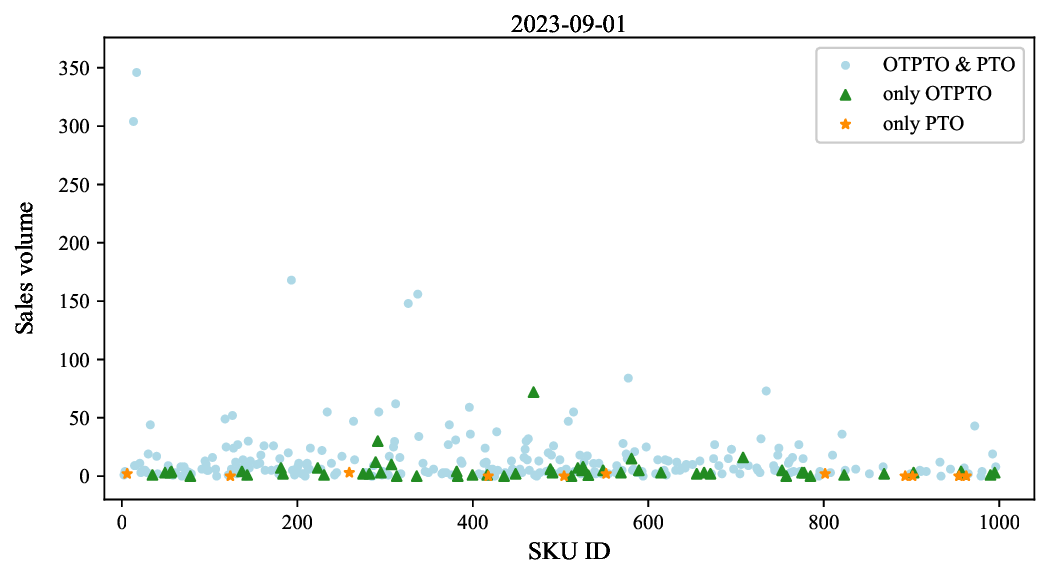}
	\caption{Product selection results of OTPTO and PTO on 2023-09-01}
	\label{fig:stock-0901}
\end{figure}

\subsection{Ablation Experiment}
To evaluate the impact of various sample, label, and feature  strategies on the OTPTO method, we conduct an ablation study. These strategies are categorized into six distinct groups: sample strategy A1, label strategy A2, decision-making feature strategy A3, sales prediction feature strategy A4, clustering feature strategy A5, and SKU-order cross-features strategy A6. 

The results presented in Table \ref{tab: re-abla} illustrates that, overall, both sample(A1) and label(A2) strategies contribute to some extent, and feature strategies exert a considerable impact, validating the significant role of these strategies. At a more granular level, the sales prediction feature(A4) is key, emphasizing that only SKUs with active sales need inventory. This is followed by decision-making features(A3), indicating that historical stocking strategies offer valuable insights for future decisions.
\begin{table}[!htbp] 
    \centering
    \caption{Results of Ablation Experiment}
    \label{tab: re-abla}
    \renewcommand{\arraystretch}{1.2}
    \begin{tabular}{cccccccc}
    \toprule
    \textbf{Date} & \multicolumn{7}{c}{\textbf{Full order fulfillment rate}}  \\ \hline 
    ~ & OTPTO & A1 & A2 & A3 & A4 & A5 & A6  \\ \cline{2-8}
    2023-09-01 & 72.11\si{\percent} & 71.05\si{\percent} & 70.69\si{\percent} & 69.89\si{\percent} & 65.28\si{\percent} & 73.09\si{\percent} & 70.24\si{\percent} \\
    2023-09-02 & 58.21\si{\percent} & 57.56\si{\percent} & 56.34\si{\percent} & 55.85\si{\percent} & 50.24\si{\percent} & 58.53\si{\percent} & 57.24\si{\percent} \\
    2023-09-03 & 64.02\si{\percent} & 63.33\si{\percent} & 63.33\si{\percent} & 63.11\si{\percent} & 56.21\si{\percent} & 62.73\si{\percent} & 62.80\si{\percent} \\
    2023-09-04 & 66.14\si{\percent} & 65.88\si{\percent} & 64.99\si{\percent} & 64.91\si{\percent} & 57.17\si{\percent} & 66.57\si{\percent} & 65.88\si{\percent} \\
    2023-09-05 & 69.32\si{\percent} & 69.22\si{\percent} & 69.22\si{\percent} & 67.75\si{\percent} & 59.59\si{\percent} & 69.03\si{\percent} & 69.91\si{\percent} \\
    2023-09-06 & 61.91\si{\percent} & 60.20\si{\percent} & 60.72\si{\percent} & 60.80\si{\percent} & 53.03\si{\percent} & 61.14\si{\percent} & 60.72\si{\percent} \\
    2023-09-07 & 69.66\si{\percent} & 68.54\si{\percent} & 69.38\si{\percent} & 68.35\si{\percent} & 61.05\si{\percent} & 68.63\si{\percent} & 70.22\si{\percent} \\
    Avg & 65.91\si{\percent} & 65.11\si{\percent} & 64.95\si{\percent} & 64.38\si{\percent} & 57.51\si{\percent} & 65.67\si{\percent} & 65.28\si{\percent} \\
    \midrule
    \textbf{Diff} & - & \textbf{0.80\si{\percent}} & \textbf{0.96\si{\percent}} & \textbf{1.53\si{\percent}} & \textbf{8.40\si{\percent}} & \textbf{0.24\si{\percent}} & \textbf{0.63\si{\percent}} \\
    \bottomrule
    \end{tabular}
\end{table}

 \subsection{Robustness Analysis} 
\begin{figure}[htbp]
	\centering
	\includegraphics[width=0.9\columnwidth]{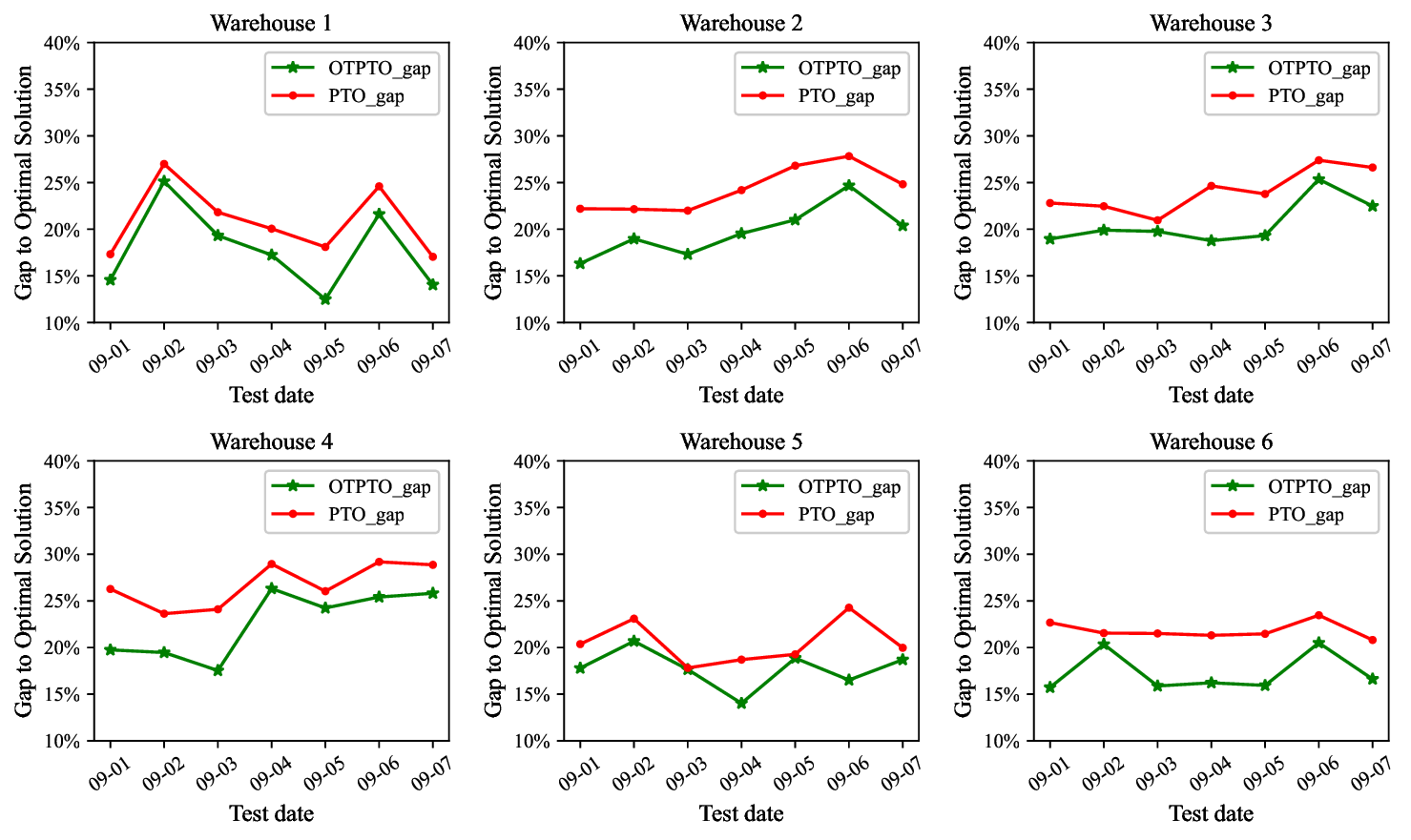}
	\caption{Gap between the OTPTO/PTO method and the optimal solution}
	\label{fig:robust}
\end{figure}

To eliminate the impact of a particular front-end warehouse, we have conducted numerical experiments using real data from other five front-end warehouses. The experimental results show that the gap between the full order fulfillment rate of the OTPTO method and the optimal solution is significantly smaller than that of the PTO method, demonstrating the robustness of the OTPTO method's performance(see Figure \ref{fig:robust}).

\section{Conclusion}
Aligning with JD.com’s business principle of “customer first”, this paper addresses the front-end warehouse inventory problem by modeling it as a multi-task problem that involves joint product selection and inventory optimization. Our objective is to increase consumer satisfaction by enhancing the full order fulfillment rate. To achieve this, we propose an innovative Optimize-then-Predict-then-Optimize (OTPTO) method. By determining historically optimal inventory plans during the first optimization phase, it is used as the features and labels in prediction phase to forecast the types and quantities of inventory goods in the future. After the second optimization phase, the final results are obtained. This method offers significant advantages over the traditional Predict-then-Optimize (PTO) approach by resolving the inconsistency between prediction goals (sales volume) and decision goals (inventory goods and quantities). To further address challenges such as sample consistency inherent in the OTPTO method, we propose a set of enhanced strategies for sampling, labeling, and feature engineering, which ensure the prediction models are more accurate and reliable. Through numerical experiments based on the transaction data from 7Fresh, JD.com's fresh food supermarket, we validate the effectiveness and robustness of our OTPTO method. The results demonstrate that this method significantly improves the full order fulfillment rate and effectively addresses product selection and inventory issues in front-end warehouses, under the constraints of limited capacity and daily fluctuations in inventory types and quantities. Future research could focus on resolving sample inconsistencies caused by the OTPTO approach and exploring its applicability in a wide range of scenarios.

\newpage
{\small
\bibliographystyle{unsrt}
\bibliography{references}
}

\newpage
\appendix

\section{appendix}
\subsection{Algorithms}
\label{alg:Algorithms}

\begin{algorithm}[htbp]
\caption{$y_{i}^\star$ Generation and Smoothing Algorithm (Part 1)}
\label{alg:Algo1}
\begin{algorithmic}[1]
      \State {\bf Input} order data, $K$, $N$, $B$, $\lambda$, $\mu$, $\gamma$.
      \State {\bf Output} $y_i^\star$.
      \State \textbf{\# 1 Define K-Means features:}
      \State $sale\, qtty_{ti}:$ The sales volume of SKU $i$ on day $t$.
      \State $ord\, cnt_{ti}:$ The cover order count of SKU $i$ on day $t$.
      \State $gmv_{ti}:$ The gross merchandise volume of SKU $i$ on day $t$.
      \State $sku\, ord\, sale\, qtty\, mean_{ti}:$ The average quantity of sales in orders associated with SKU $i$ on day $t$.
      \State $sku\, ord\, sku\, mean_{ti}:$ The average number of SKU in orders associated with SKU $i$ on day $t$.

      \State \textbf{\# 2 $y_{ti}^\star$ Generation:}
      \For{historical period $t=1$ \textbf{to} $T$}
          \State OM1 $\gets$ Create the OM1 model with objective (\ref{eq:om-obj1&obj2}) based on the given input data.
           \State $x_{ti}^\star, y_{ti}^\star$ $\gets$ Solve the OM1 using commercial solver.
      \EndFor

        \State \textbf{\# 3 $y_{ti}^\star$ Smoothing:}
        \State \textbf{\# 3.1 $y_{ti}^\star$ Cross Sectional Smoothing:}
        \For{historical period $t=1$ \textbf{to} $T$}
            \For{SKU $i=1$ \textbf{to} $m$}
                \State Calculate K-Means features.
            \EndFor
            \State $cs\,data_t$ $\gets$ Build K-Means data.
            \State $cs\,data_t$ $\gets$ MinMaxNorm($cs\,data_t$).
            \State $cs\, results_t$ = KMeans($cs\,data_t$, \, $clusters=\lambda$).
            \For{cluster $c$ in $cs\, results_t$}
                \State $ncs$ $\gets$ The number of SKU in cluster.
                \State Let $\alpha=0$.
                
                \For{SKU $i=1$ \textbf{to}  $ncs$}
                    \If{$y_{ti}^\star = 1$}
                        \State $\alpha\, +=1$.
                    \EndIf
                \EndFor
                
                \For{SKU $i=1$ \textbf{to}  $ncs$}
                    \State Let $ycs_{ti} = y_{ti}^\star$
                    \If{$\alpha / ncs > \mu$}
                        \State $ycs_{ti} = 1$.
                    \EndIf
                \EndFor
            \EndFor
        \EndFor
      \algstore{myalgorithm} 
\end{algorithmic}
\end{algorithm}

\begin{algorithm}[t]
\ContinuedFloat 
\caption{$y_{i}^\star$ Generation and Smoothing Algorithm (Part 2)}
\begin{algorithmic}[1]
\algrestore{myalgorithm} 
        \State \textbf{\# 3.2 Time Series Smoothing:}
        \For{SKU $i=1$ \textbf{to} $m$}
            \State Let $nts=0, \beta=0$.
            \For{historical period $t=1$ \textbf{to} $T$}
                \If{$sale\,qtty_{ti} > 0$}
                    \State $nts \,+= 1$.
                    \If{$y_{ti}^\star = 1$}
                        \State $\beta \, += 1$
                    \EndIf
                \EndIf
            \EndFor
            \For{historical period $t=1$ \textbf{to} $T$}
            \State Let $yts_{ti} = y_{ti}^\star$
                \If{$\beta/nts > \gamma$}
                    \State $yts_{ti} = 1$
                \EndIf
            \EndFor
        \EndFor
        \State \textbf{\# 3.3 Smoothing Merge:}
        \For{historical period $t=1$ \textbf{to} $T$}
            \For{SKU $i=1$ \textbf{to} $m$}
                \If{$ycs_{ti}=1$ and $yts_{ti}=1$}
                    \State $y_{ti}^\star = 1$.
                \EndIf
            \EndFor
        \EndFor
\end{algorithmic}
\end{algorithm}


\begin{algorithm}[t]
\caption{Results Post-Processing Algorithms.}
\label{alg:Algo2}
    \begin{algorithmic}[1]
        \State {\bf Input} $\hat{y}_{ti}, \hat{x}_{ti}, \hat{q}_{ti}, K, N, B, \text{algo type}$.
        \State {\bf Output} final result.
        \State $\hat{y}_{ti} \gets$ The Probability of SKU $i$ being selected from PM1 results.
        \State $\hat{x}_{ti} \gets$ The Quantity of SKU $i$ inventory from PM2 results.
        \State $\hat{q}_{ti} \gets$ The Sales prediction value of SKU $i$ on day $t$.
        \For{future period $t = 1$ \textbf{to} $T$}
            \State \textbf{\# 1 Select SKUs to be stocked:}
            \If{algo type = 'OTPTO'}
                \State $sorted\,skus \gets$ Sort SKUs in descending order according to $\hat{y}_{ti}$.
                \State $inventory\,skus \gets$ Select top $K$ SKUs from $sorted\,skus$.
                \State $sorted\,inventory\,skus \gets$ Sort selected SKUs in descending order according to $\hat{x}_{ti}$.
                \For{SKU $i$ in $sorted \, inventory \, skus$}
                    \State $qtty_{ti} = \max(B, \min(\hat{x}_{ti}, \hat{q}_{ti}))$
                \EndFor
            \ElsIf{algo type = 'PTO'}
                \State $sorted\, skus \gets $ Sort SKUs in descending order according to $\hat{y}_{ti}$.
                \State $sorted\,inventory\,skus \gets$ Select top $K$ SKUs from $sorted\,skus$.
                \For{SKU $i$ in $sorted\, inventory\, skus$}
                    \State $qtty_{ti} = \max(B, \hat{q}_{ti})$.
                \EndFor
            \EndIf
            \State \textbf{\# 2.Greedy Search:}
            \State Let $\alpha = N/\sum_{i\in sorted\,inventory\,skus} qtty_{ti}$.
            \State Let $final\, result$ = \{\,\}\, $inventory\,qtty_t=0$.
            \For{SKU $i$ in $sorted \, inventory\, skus$}
                \State $qtty_{ti} = round(qtty_{ti} * \alpha,0) $.
                \If{$qtty_{ti} + \, inventory\, qtty_{t} \leq N$}
                    \State $inventory\, qtty_t\, += qtty_{ti}$.
                    \State $final\, result \,\mathrm{append}\,(t, sku_{ti}, qtty_{ti})$.
                \Else
                    \State \textbf{break}.
                \EndIf  
            \EndFor
        \EndFor
\end{algorithmic}
\end{algorithm}

\newpage 
\onecolumn 

\subsection{Tables}
\label{tab:Tables}
\begin{table*}[!ht]
    \centering
    \caption{Parameters of PM1/PM2}
    \label{tab:PM1-PM2-param}
    \begin{tabular}{ccc}
    \toprule
        Parameters Name & PM1 Value & PM2 Value  \\ \hline
        boosting\_type & 'gbdt' & 'gbdt'  \\ 
        objective & 'binary' & 'regression'  \\ 
        metric & 'auc' & 'rmse'  \\ 
        reg\_alpha & 0 & 0.1  \\ 
        reg\_lambda & 0 & 0.1  \\ 
        subsample & 0.8 & 0.8  \\ 
        subsample\_freq & 1 & 1  \\ 
        learning\_rate & 0.05 & 0.1  \\ 
        num\_leaves & 31 & 31  \\ 
        max\_depth & 5 & 5  \\ 
        min\_child\_samples & 5 & 5  \\ 
        colsample\_bytree & 0.8 & 0.8  \\ 
        n\_estimators & 600 & 600  \\ 
        eval\_metric & 'auc' & 'rmse'  \\ 
        early\_stopping\_rounds & 50 & 50  \\ 
        \bottomrule
    \end{tabular}
\end{table*}

\begin{table*}[!ht]
    \centering
    \caption{Results of Inventory Resource Allocation}
    \label{tab:re-inv}
    \renewcommand{\arraystretch}{1.2}
    \begin{tabular}{ccccccccc}
    \toprule
        \textbf{Date} & \multicolumn{2}{c}{\textbf{Number of SKUs}}  & \multicolumn{2}{c}{\textbf{Total Inventory}}  & \multicolumn{2}{c}{\textbf{min/max inventory}}  & \multicolumn{2}{c}{\textbf{avg/median inventory}}  \\ \cline{2-9}
        ~ & \textbf{OTPTO} & \textbf{PTO} & \textbf{OTPTO} & \textbf{PTO} & \textbf{OTPTO} & \textbf{PTO} & \textbf{OTPTO} & \textbf{PTO} \\ \hline
        2023-09-01 & 347 & 299 & 8999.4 & 8990.8 & 10/478.7 & 10/505.9 & 25.9/15 & 30.1/16  \\ 
        2023-09-02 & 347 & 298 & 8999.4 & 8990.7 & 10/510.5 & 10/682.1 & 25.9/15 & 30.1/15  \\ 
        2023-09-03 & 347 & 299 & 8999.1 & 8992.3 & 10/392.0 & 10/558.2 & 25.9/15 & 30.1/15  \\ 
        2023-09-04 & 346 & 297 & 8992.4 & 8991.2 & 10/398.0 & 10/499.5 & 26.0/16 & 30.3/15  \\ 
        2023-09-05 & 347 & 299 & 8994.6 & 8991.6 & 10/403.6 & 10/514.0 & 25.9/16 & 30.1/15  \\ 
        2023-09-06 & 347 & 301 & 8994.1 & 8999.9 & 10/545.8 & 10/525.1 & 25.9/15 & 29.9/14  \\ 
        2023-09-07 & 349 & 300 & 8997.4 & 8992.4 & 10/401.8 & 10/496.9 & 25.8/16 & 30.0/15  \\ 
        \bottomrule
    \end{tabular}
\end{table*}
\end{document}